\documentclass{article}
\usepackage{spconf,amsmath,graphicx,setspace}
\usepackage{amssymb}
\usepackage{algorithm}
\usepackage{algorithmic}
\usepackage{textcomp}
\usepackage{subfigure}
\usepackage{tipa}
\usepackage{cite} 
\usepackage{url}
\usepackage{float}
\usepackage{arydshln}
\usepackage{booktabs}
\usepackage{multirow}
\usepackage{makecell}
\usepackage{soul}
\usepackage{xcolor}

\makeatletter 
\renewcommand{\@thesubfigure}{\hskip\subfiglabelskip}
\makeatother

\title{PERSONALIZED FEDERATED LEARNING FOR EGOCENTRIC VIDEO GAZE \\ ESTIMATION WITH COMPREHENSIVE PARAMETER FREEZING}
\name{Yuhu Feng $^{\dagger}$ \qquad Keisuke Maeda $^{\dagger\dagger}$ \qquad Takahiro Ogawa $^{\dagger\dagger\dagger}$ \qquad Miki Haseyama$^{\dagger\dagger\dagger}$ 
}
\address{$^{\dagger}$ Graduate School of Information Science and Technology,
    Hokkaido University, Japan \\
    $^{\dagger\dagger}$ Data-Driven Interdisciplinary Research Emergence Department, Hokkaido University, Japan \\
    $^{\dagger\dagger\dagger}$ Faculty of Information Science and Technology, 
    Hokkaido University, Japan \\
 	E-mail: \{feng, maeda, ogawa, mhaseyama\}@lmd.ist.hokudai.ac.jp
    }

\begin{document}
\ninept
\maketitle
%
%
\begin{abstract}
Egocentric video gaze estimation requires models to capture individual gaze patterns while adapting to diverse user data.
%
We propose a Personalized Federated Learning (PFL) framework with Comprehensive Parameters Freezing (FedCPF) to improve gaze estimation in distributed environments.
Our approach leverages a transformer-based architecture, integrating it into a PFL framework where only the most significant parameters—those exhibiting the highest rate of change during training—are selected and frozen for personalization in client models. 
Through extensive experimentation on the EGTEA Gaze+ and Ego4D datasets, we demonstrate that FedCPF significantly outperforms previously reported federated learning methods, achieving superior recall, precision, and F1-score. 
These results confirm the effectiveness of our comprehensive parameters freezing strategy in enhancing model personalization, making FedCPF a promising approach for tasks requiring both adaptability and accuracy in federated learning settings.
\end{abstract}
\begin{keywords}
personalized federated learning, parameter freezing, gaze estimation, egocentric video.
\end{keywords}

\vspace{-3mm}
\section{Introduction}
\label{sec:intro}
\vspace{-2mm}
Egocentric videos offer great potential for gaze estimation \cite{betancourt2015evolution} in AR, VR, and assistive technologies, but raise privacy concerns due to sensitive personal data \cite{dimiccoli2018mitigating}.
Traditional methods \cite{li2013learning, huang2018predicting} typically rely on a centralized training approach, where data from various users are aggregated into a large dataset and uploaded to a central server for training.
This centralized data collection and transmission process poses a significant risk of privacy breaches \cite{jia2019automatic}. 
Additionally, models trained on aggregated data, while possessing robust generalization capabilities, often fail to capture the subtle differences between individual user data, resulting in a lack of personalization for different users \cite{arivazhagan2019federated}.

Federated Learning (FL) allows distributed training without centralizing data but faces performance issues due to client data variation \cite{collins2021exploiting}.
%
Personalized Federated Learning (PFL) enhances FL by tailoring models to individual clients' data distributions \cite{tan2022towards}.
This approach typically involves sharing a global model and locally adjusting it for each client, thereby enhancing model performance while maintaining the advantages of data privacy and collaborative learning \cite{arivazhagan2019federated}.
Recently, various PFL methods have been proposed. 
For example, methods in \cite{arivazhagan2019federated, collins2021exploiting} focus on fine-tuning the parameters of the prediction head in client models for personalization, while methods in \cite{liang2020think, oh2021fedbabu} consider personalizing the feature extractor of client models. 
However, recent studies \cite{frankle2018lottery, frankle2020linear, renda2020comparing} indicate that the importance of parameters for prediction can vary significantly even within the same layer. 
The coarse-grained layer-wise selection of personalized sub-networks may not fully balance the knowledge sharing among clients and the personalization of each client. 
To address this problem, FedSelect \cite{tamirisa2024fedselect}, which selects parameters for personalization from a global perspective rather than focusing on a specific layer, was proposed. 
It assumes that each client model contains a sub-network crucial for personalization to the local data distribution \cite{li2020lotteryfl}, and personalizing this sub-network can achieve the optimal balance between global knowledge sharing and local personalization. 
%
Parameters with the highest rate of change are frozen for personalization, while others are aggregated into the global model.

However, the parameter selection function in FedSelect still has its limitations. 
It focuses on the parameters with the highest rate of change in every communication round, neglecting the impact of parameters with slightly lower change rates on personalization \cite{ho2014effects}. 
Moreover, it overlooks the fact that some parameters may undergo significant changes only in certain updates but remain stable in subsequent ones \cite{luong1996fundamental}, failing to consider the cumulative impact of these parameters on personalization from a global perspective. 
As a result, important parameters may not be selected, limiting the personalization performance.

In this paper, we propose a PFL framework with Comprehensive Parameters Freezing (FedCPF), which accounts for the rate of parameter change over multiple training iterations.
Specifically, rather than using the rate of parameter change in a single communication round as the sole criterion for selecting personalized parameters, we compute the average rate of parameter change over multiple updates.
%
This comprehensive consideration of parameter changes during the training process allows for a more accurate selection of personalized parameters, which are then frozen in the client models, thereby enhancing the performance of the model within the PFL. 
Experimental results show that our framework outperforms other FL methods, achieving higher evaluation metrics and demonstrating the effectiveness of its parameter freezing approach.

\vspace{-3mm}
\section{Gaze Estimation via FedCPF}
\label{sec:method}
\vspace{-2mm}
In this section, we present our approach to egocentric video gaze estimation, employing a comprehensive parameter freezing strategy within a PFL framework. 
Specifically, we leverage the Global-Local Correlation (GLC) module \cite{lai2024eye}, a transformer-based architecture, to predict user gaze in egocentric videos, which serves as the baseline within the PFL framework.
The GLC module enhances video representation learning by capturing additional correlations between video frame regions via self-attention mechanisms. 
Additionally, the parameter matrices within the self-attention layers are individually updated and trained for each client in the PFL framework. 
%
%
We categorize the parameters in the client models into two distinct groups: ``personalized" and ``global." 
The personalized parameters are those that are most critical for capturing the characteristics of the local data and are thus frozen within the client during the training process. 
The unfrozen parameters are designated as global parameters, which are then aggregated into the global model. 

\begin{figure}[t]
    \centering
    \includegraphics [clip,scale=0.4]{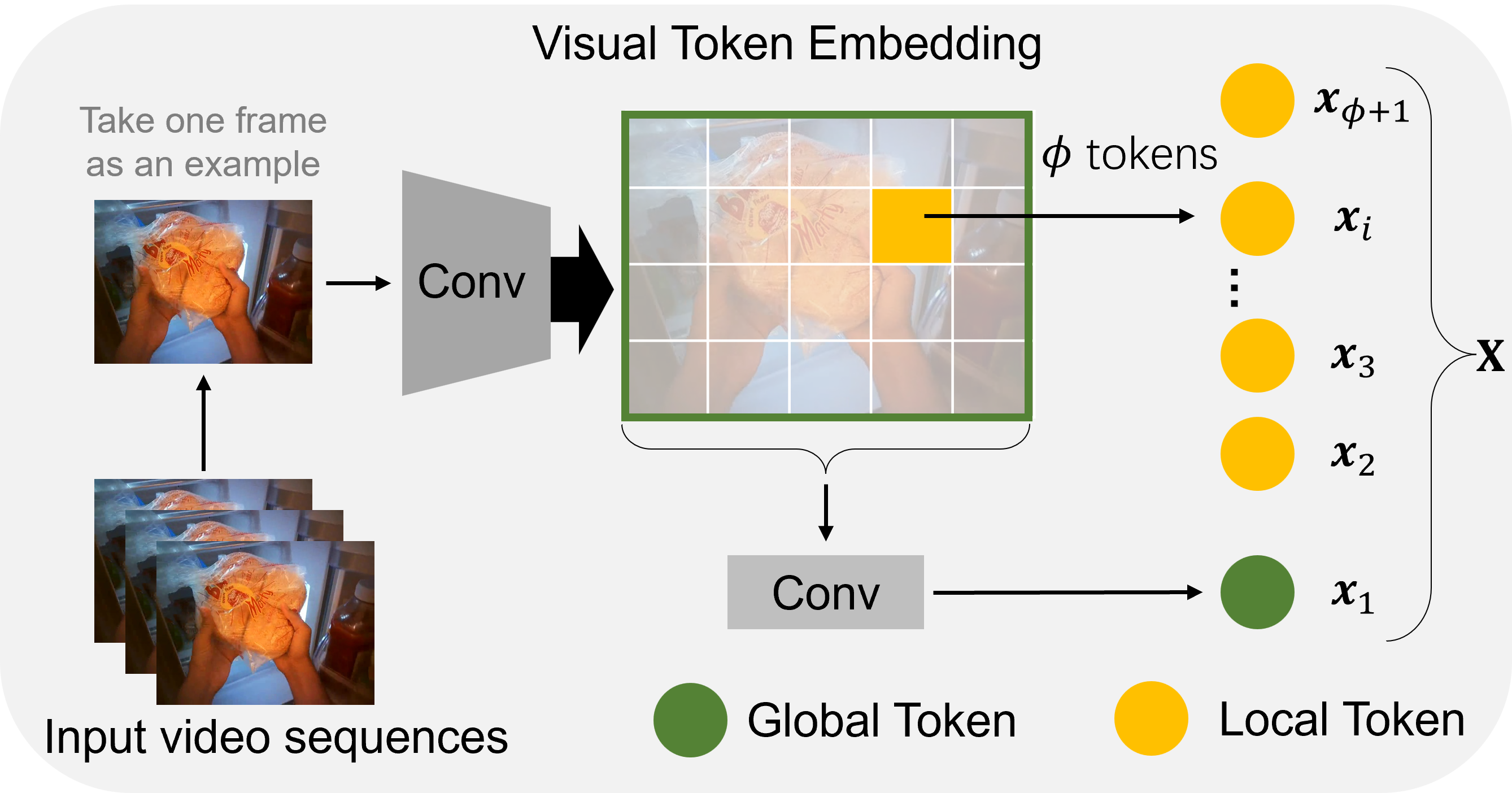}
    \vspace{-3mm}
    \caption{An overview of the visual token embedding process in the GLC module. Each input video frame is divided into patches, processed by convolution, and embedded into global and local tokens for further analysis. The global token ($\textbf{\textit{x}}_1$) and local tokens ($\textbf{\textit{x}}_2$ to $\textbf{\textit{x}}_{\phi+1}$) are generated for subsequent processing.}
    \vspace{-2mm}
    \label{GLC Visual Token Embedding}
\end{figure}

\vspace{-3mm}
\subsection{Global-Local Correlation Module}
\vspace{-2mm}
The GLC module operates as a transformer-based framework for egocentric video gaze estimation, performing as the core model on both the client and server sides. 
As shown in Fig. \ref{GLC Visual Token Embedding}, the input egocentric video sequence is divided into multiple patches, which are processed into visual tokens within the embedding part of the GLC module.
Each patch is then independently embedded into a vector space through a convolutional layer.
%
This embedding process produces a set of tokens, denoted as $\textbf{\textit{x}}$, representing local features, with a total number of $\phi$ tokens. 
These tokens are subsequently transformed into an embedding matrix $\textbf{\textit{M}}$. 
In each self-attention layer within the encoder block, the query, key, and value matrices are respectively computed as $\textbf{Q} = \textbf{\textit{M}}\textbf{\textit{W}}^Q$, $\textbf{K} = \textbf{\textit{M}}\textbf{\textit{W}}^K$, and $\textbf{V} = \textbf{\textit{M}}\textbf{\textit{W}}^V$. 
We collectively denote the weight matrices as $\textbf{\textit{W}} = [\textbf{\textit{W}}^Q, \textbf{\textit{W}}^K, \textbf{\textit{W}}^V]$.
The self-attention mechanism is subsequently applied as follows:
\begin{equation}
    {\rm Atten}(\mathbf{Q}, \mathbf{K}, \mathbf{V}) = {\rm Act}(\mathbf{Q}\mathbf{K}^T / \sqrt{d})\mathbf{V},
\end{equation}
where $d$ represents the number of columns in the $\mathbf{Q}$, $\mathbf{K}$, and $\mathbf{V}$ matrices, and $\rm {Act}(\cdot)$ refers to the softmax activation function.
%
The GLC module processes $\phi$ local tokens and integrates an additional global token via convolutional operations, which are subsequently input into a transformer encoder consisting of self-attention blocks. 
For a transformer with $A$ blocks, the total input to the encoder at the $j$-th block is defined as $\mathbf{X}^j = (\textit{x}_1^j, \textit{x}_2^j, \dots, \textit{x}_{\phi+1}^j)$, where $j \in \{1, 2, \dots, A\}$. 
The global token is represented by the first row vector of $\mathbf{X}$, i.e., $\textbf{\textit{x}}_1$, consistent with the approach described in a previous study \cite{lai2024eye}.
To explicitly capture the contextual relationships between the global token and the local tokens derived from the egocentric video, we compute the correlation between each local token and the global token, denoted as Correlation($\textbf{\textit{x}}_i$, $\textbf{\textit{x}}_1$), along with its self-correlation, Correlation($\textbf{\textit{x}}_i$, $\textbf{\textit{x}}_i$). 
These correlation scores are subsequently normalized via the softmax function to fine-tune their corresponding weights. 
Furthermore, we introduce a suppression matrix $\mathbf{S}^{(\phi+1)\times(\phi+1)}$ to reduce the influence of the correlations among the other tokens. This matrix is defined as follows:
\begin{equation}
    \mathbf{S}^{(\phi+1)\times(\phi+1)} = \left\{\begin{matrix}
    \begin{aligned}
        0,&\text{ if }i=j \text{ or }j=1 ,\\
        \lambda, &\text{ otherwise.} \notag
    \end{aligned}
    \end{matrix}\right.
\end{equation}
In this equation, the diagonal elements and the elements in the first column of the matrix $\mathbf{S}$ are set to zero, while the remaining elements are assigned a large value $\lambda$, following the empirical method used in \cite{liu2021swin}. 
Finally, the self-attention function within the GLC module can be expressed as follows:
\begin{equation}
    {\rm Atten}(\mathbf{Q}, \mathbf{K}, \mathbf{V}) = {\rm Act}((\mathbf{Q}\mathbf{K}^T-\mathbf{S}) /\sqrt d)\mathbf{V}.
\end{equation}

\begin{figure*}[t]
    \centering
    \includegraphics[clip, scale=0.3]{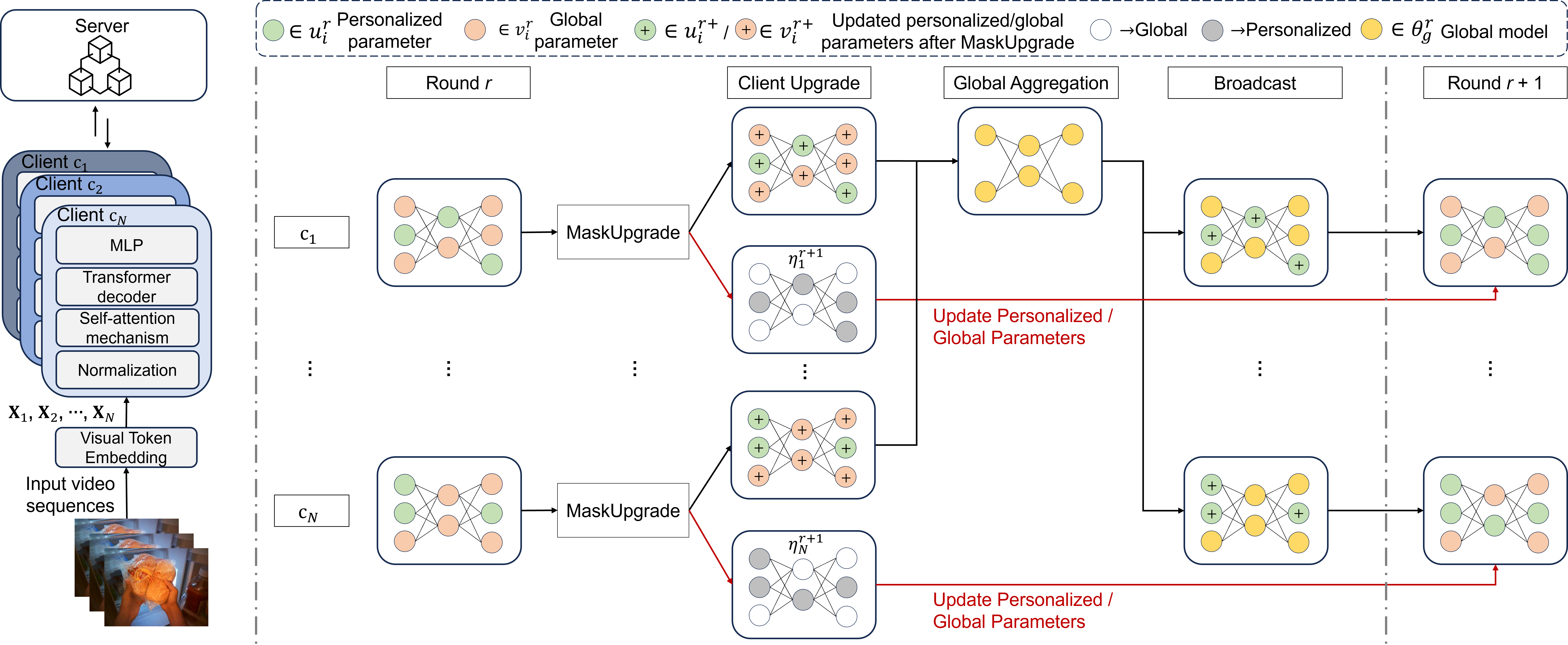}
    \caption{An overview of FedCPF algorithm. Input video sequences are embedded into tokens, processed by client models using a transformer with a Global-Local Correlation mechanism. Personalized parameters are frozen locally, while global parameters are aggregated to update the global model. $u_i^r$ and $v_i^r$ are the personalized$/$global parameters, while $u_i^{r+}$ and $v_i^{r+}$ denote the updated personalized/global parameters after MaskUpgrade, respectively. And $\theta_g^r$ represents the current global model.}
    \vspace{-3mm}
    \label{overview}
\end{figure*}

\setlength{\textfloatsep}{6pt} 
\begin{algorithm}[t]
\caption{FedCPF}
\label{alg1}
\begin{algorithmic}[1]
    \STATE Input: Client $\rm{C} = \{\rm{c}_1, \rm{c}_2, ..., \rm{c}_\textit{N}\}$, communication rounds $\textit{R}$, initial foundation model $\theta_g^0$
    \STATE Set all client models ${\theta_i^0}$ for $i \in \{1, N\}$ with $\theta_g^0$ 
    \STATE Set all client binary masks ${\eta_i^0}$ for $i \in \{1, N\}$ with 0
    \FOR{each round $r$ in 0, 1, 2, ..., \textit{R}-1}
        \FOR{each client $\rm{c}_\textit{i} \in \rm{C}$}
        \STATE $u_i^r \leftarrow \theta_i^r \odot \eta_i^r$, $v_i^r \leftarrow \theta_i^r \odot \neg \eta_i^r$
        \STATE $u_i^{r+}, v_i^{r+}, \eta_i^{r+1} \leftarrow$ \text{MaskUpgrade($u_i^r, v_i^r$)} 
        \STATE $\theta_g^r \leftarrow 0$, $\zeta^r \leftarrow 0$
        \ENDFOR
        \FOR{each client $\rm{c}_\textit{i} \in \rm{C}$}
        \STATE $\theta_g^r \leftarrow \{(\theta_g^r \odot \neg \eta_i^r) + v_i^{r+}\} \odot \neg \eta_i^r$
        \STATE $\zeta^r \leftarrow \{(\zeta^r \odot \neg \eta_i^r) + \neg \eta_i^r\} \odot \neg \eta_i^r$
            \ENDFOR
        \STATE $\eta_g^r \leftarrow$ Binary mask for $\zeta^r \neq 0$
        \STATE $\theta_g^r \leftarrow \frac{\theta_g^r \odot \eta_g^r}{\zeta^r \odot \eta_g^r} \odot \eta_g^r$
        \FOR{each client $\rm{c}_\textit{i} \in \rm{C}$}
        \STATE $\theta_i^{r+1} \leftarrow \theta_g^r \odot \neg \eta_i^r + u_i^{r+} \odot \eta_i^r$
        \ENDFOR
    \ENDFOR
\end{algorithmic}
\end{algorithm}

\vspace{-3mm}
\subsection{Personalized Federated Learning with Comprehensive Parameters Freezing}
\vspace{-2mm}
We introduce our Personalized Federated Learning with Comprehensive Parameters Freezing (FedCPF), a framework designed to select and freeze parameters that exhibit the most significant changes during training, enabling effective personalization in client models.
%
This process is designed to comprehensively select only those parameters that exhibit the most significant changes during training as personalized parameters for client models in PFL, and then freeze these parameters in the client models.
As shown in Fig. \ref{overview}, we assume that there are $N$ clients. 
For each client $\rm{c}_\textit{i} \in \rm{C}$, training is conducted on the local dataset $D_i = \{\mathbf{X}_i, \mathbf{Y}_i\}$ ($i \in N$), where $\mathbf{X}_i$ denotes the total input of all visual tokens for each client, and $\mathbf{Y}_i$ represents the gaze fixation ground truth.
%
%
$D = \bigcup_{i \in N} D_i$ represents the total dataset of size $\nu = \sum_{i=1}^{N} \phi_i$, and $f_i(\theta_i; \cdot)$ denotes the model for client $\rm{c}_\textit{i}$, where $\theta_i$ are the model parameters.
%
%
Finally, the objective function is defined as
\begin{equation}
    \text{arg} \min_\Theta  \sum_{i=1}^{N} \frac{\phi_i}{\nu} l(f_i(\theta_i; \mathbf{X}_i), \mathbf{Y}_i),
    \label{initial loss function}
\end{equation}
where $\Theta = \{\theta_i\}_{i=1}^N$, and $l(\cdot, \cdot)$ is the cross-entropy loss function applied to all clients. 
The objective of this formulation is to determine the personalized model parameters $\theta_i$ for each client $\rm{c}_\textit{i}$.

To facilitate the explanation of the comprehensive parameters freezing process in Algorithms \ref{alg1} and \ref{alg2}, we introduce the following definition. 
%
%
%
Parameters $\theta_i$ in client models are split into two partitions: $u_i$ for personalization (frozen in local models) and $v_i$ for aggregation into the global model $\theta_g$.
Furthermore, for a parameter matrix $\theta$ and a binary mask matrix $\eta$, a value of 1 in $\eta$ indicates that the corresponding position in $\theta$ retains its value (weight), while a value of 0 in $\eta$ sets the corresponding position in $\theta$ to zero. 
We denote this operation as $\theta \odot \eta$. Additionally, we use $\neg$ to represent the complement operation on a binary matrix.

FedCPF takes the following as input: the set of clients $\rm{C}$, the number of communication rounds $R$, the freezing rate $p$, and the personalization rate $\rho$. 
At the start of PFL, all client models are initialized with the same foundational model parameters $\theta_g^0$, represented by binary client masks set to zero matrices ($\eta_i^0 = 0$) applied to each client model $\theta_i^0$.
%
%
Following the method from the previous research \cite{tamirisa2024fedselect}, in the current communication round $r$, we first update each client’s current global parameters $v_i^r$ and personalized parameters $u_i^r$; these are identified using the current mask $\eta_i^r$ as an index into $\theta_i^r$. 
Additionally, $v_i^{r+}$ and $u_i^{r+}$ denote the updated personalized$/$global parameters after MaskUpgrade (Algorithm \ref{alg2}), respectively.
%
%
The parameters exhibiting the highest $p\%$ average change rate are retained as personalized parameters in the client model, while the remaining parameters are aggregated into the global model as global parameters.
This process is denoted by setting the corresponding indices for $u_i^r$ in $\eta_i^r$ to 1. 
%
%
%
This formula divides $\theta_i^r$ into two parts: $\theta_i^r \odot \eta_i^r$ for freezing, and $\theta_i^r \odot \neg \eta_i^r$ for global aggregation.

Once each client updates its model, the global parameters $v_i$ are aggregated. 
In typical PFL settings, careful consideration is required when averaging $v_i$ across clients, as the mask $\eta_i^r$ is likely heterogeneous. 
This means the locations of global parameters (where $\eta_i$ equals 0) may differ across clients.
%
%
%
In FedCPF, global averaging for a specific index in the current global model $\theta_g^r$ is only applied to the parameters $v_i$ from clients where the corresponding entry in $\eta_i^r$ is 0. 
We introduce $\zeta^r$ to monitor the number of clients contributing to each global parameter in $\theta_g^r$ on an element-wise basis. 
This setup allows different subsets of clients $\rm{c}_\textit{i} \in \rm{C}$ to contribute to various global parameters in $\theta_g^r$.
%
%
%
%


\setlength{\textfloatsep}{6pt} 
\begin{algorithm}[t]
\caption{MaskUpgrade($u_i^r, v_i^r$)}
\label{alg2}
    \begin{algorithmic}[1]
    \STATE Input global/personalized parameters $u_i^r$ and $v_i^r$, local epochs $E$, freezing rate $p$, personalization rate $\rho$, accuracy threshold $acc$, control parameters $k$ and $r_{th}$
    \STATE $k \leftarrow 0$
    
    \FOR{$e = 0, 1, ..., \textit{E}-1$}
        \STATE $\theta_i^{r, e+1} \leftarrow \text{arg } \min_\Theta  \sum_{i=1}^{N} \frac{\phi_i}{\nu} l(f_i(\theta_i; \mathbf{X}_i), \mathbf{Y}_i)$
        \ENDFOR
    \STATE $u_i^{r+} \leftarrow \theta_i^{r, \textit{E}} \odot \eta_i^r$, $v_i^{r+} \leftarrow \theta_i^{r, \textit{E}} \odot \neg \eta_i^r$
    \IF{sparsity of $\neg \eta_i^r < \rho$}
        \IF{the the client model accuracy $\geq k \cdot acc$}
            \STATE $r_{th} \leftarrow r$
            \STATE $k \leftarrow k+1$
            \ENDIF
        \STATE $\Delta_{v_i^r} \leftarrow \vert v_i^{r+} - v_i^r \vert$
        \STATE $\Delta_{v_i^r}^{\rm{avg}} \leftarrow \frac {\sum_{r = r_{\rm{th}}}^{r_{\rm{cu}}} \Delta v_i^r} {r_{\rm{cu}} - r_{\rm{th}}}$
        \STATE $\eta_i^{r+1} \leftarrow$ binary mask for highest $p\%$ value in $\Delta_{v_i^r}^{\rm{avg}}$
        \STATE $\eta_i^{r+1} \leftarrow \eta_i^{r+1} \cup \eta_i^r$
    \ELSE 
        \STATE $\eta_i^{r+1} \leftarrow \eta_i^r$
    \ENDIF
    \RETURN $u_i^{r+}, v_i^{r+}, \eta_i^{r+1}$
    \end{algorithmic}
\end{algorithm}

\begin{table*}[t]
    \centering
    \vspace{-3mm}
    \caption{Quantitative evaluation results of different methods for egocentric video gaze estimation.}
    \begin{tabular}{llccc|ccc}
    \toprule
     & \textbf{Dataset} & \multicolumn{3}{c|}{\textbf{EGTEA Gaze+}} & \multicolumn{3}{c}{\textbf{Ego4D}} \\
    \cmidrule(lr){3-5} \cmidrule(lr){6-8}
    & \textbf{Method} & Recall & Precision & F1-score & Recall & Precision & F1-score \\
    \midrule
    & Local-only & 44.24 & 23.17 & 30.41 & 27.31 & 16.83 & 20.83 \\
    & FedAvg \cite{konevcny2016federated} & 27.42 & 12.95 & 17.59 & 17.62 & 10.55 & 13.20 \\
    & FedProx \cite{li2020federated} & 36.68 & 21.36 & 27.00 & 29.73 & 19.65 & 23.66 \\
    & FedPAC \cite{xu2023personalized} & 52.78 & 28.38 & 36.91 & 30.85 & 22.88 & 26.27 \\
    & FedSelect \cite{tamirisa2024fedselect} & 54.69 & 30.65 & 39.28 & 36.82 & 24.63 & 29.52 \\
    & FedCPF (random freezing) & 51.63 & 27.79 & 36.13 & 31.79 & 21.46 & 25.62 \\
    \midrule
    & \multirow{3}{*}{\parbox[l]{4cm}{FedCPF ($acc = 0.05;$ \\ $\rho = 0.5$ for EGTEA Gaze+ \\ $\rho = 0.4$ for Ego4D)}}
    & \multirow{3}{*}{\makecell{\textbf{55.85}}} & \multirow{3}{*}{\makecell{\textbf{31.74}}} & \multirow{3}{*}{\makecell{\textbf{40.48}}} & \multirow{3}{*}{\makecell{\textbf{37.10}}} & \multirow{3}{*}{\makecell{\textbf{25.07}}} & \multirow{3}{*}{\makecell{\textbf{29.92}}} \\
    & & & & & & & \\
    & & & & & & & \\
    \bottomrule
    \end{tabular}
    \vspace{-3mm}
    \label{quantitative evaluation}
\end{table*}

\begin{table}[t]
    \centering
    \caption{Result of FedCPF with different personalization rate $\rho$.}
    \begin{tabular}{llccc}
    \toprule
    \textbf{Dataset} & \textbf{$acc = 0.05$} & \textbf{Recall} & \textbf{Precision} & \textbf{F1-score} \\
    \midrule
    \multirow{7}{*}{\makecell{EGTEA \\ Gaze+}}
    & $\rho = 0.1$ & 52.40 & 29.25 & 37.56 \\
    & $\rho = 0.2$ & 52.51 & 29.17 & 37.51 \\
    & $\rho = 0.3$ & 54.79 & 30.56 & 39.24 \\
    & $\rho = 0.4$ & 55.38 & 31.06 & 39.80 \\
    & $\rho = 0.5$ & \textbf{55.85} & \textbf{31.74} & \textbf{40.48} \\
    & $\rho = 0.6$ & 54.16 & 29.49 & 38.19 \\
    & $\rho = 0.7$ & 52.46 & 28.94 & 37.32 \\
    \midrule
    \multirow{7}{*}{Ego4D} 
    & $\rho = 0.1$ & 34.19 & 22.4  & 27.07 \\
    & $\rho = 0.2$ & 34.67 & 23.43 & 27.96 \\
    & $\rho = 0.3$ & 35.96 & 23.77 & 28.62 \\
    & $\rho = 0.4$ & \textbf{37.10}  & 25.07 & \textbf{29.92} \\
    & $\rho = 0.5$ & 36.69 & \textbf{25.20}  & 29.88 \\
    & $\rho = 0.6$ & 34.86 & 22.55 & 27.39 \\
    & $\rho = 0.7$ & 33.67 & 22.38 & 26.89 \\
    \bottomrule
    \end{tabular}
    \vspace{-2mm}
    \label{different personalization rate}
\end{table}

\begin{table}[t]
    \centering
    \caption{Result of FedCPF with different accuracy threshold $acc$.}
    \begin{tabular}{llccc}
    \toprule
    \textbf{Dataset} & \textbf{$\rho = 0.5$} & \textbf{Recall} & \textbf{Precision} & \textbf{F1-score} \\
    \midrule
    \multirow{5}{*}{\makecell{EGTEA \\ Gaze+}}
    & $acc = 0.01$ & 54.61 & 30.63 & 39.25 \\
    & $acc = 0.05$ & \textbf{55.85} & \textbf{31.74} & \textbf{40.48} \\
    & $acc = 0.10$ & 54.79 & 30.56 & 39.24 \\
    & $acc = 0.15$ & 53.38 & 30.06 & 38.46 \\
    & $acc = 0.20$ & 54.81 & 30.73 & 39.38 \\
    \midrule
    \multirow{5}{*}{Ego4D}
    & $acc = 0.01$ & 36.17 & 23.33 & 28.36 \\
    & $acc = 0.05$ & \textbf{36.69} & \textbf{25.20} & \textbf{29.88} \\
    & $acc = 0.10$ & 35.13 & 24.12 & 28.60 \\
    & $acc = 0.15$ & 33.60 & 23.76 & 27.84 \\
    & $acc = 0.20$ & 33.49 & 23.89 & 27.89 \\
    \bottomrule
    \end{tabular}
    \vspace{-2mm}
    \label{different accuracy threshold}
\end{table}

\vspace{-3mm}
\subsection{MaskUpgrade}
\vspace{-2mm}
As mentioned earlier, focusing on the parameters with the highest rate of change in every communication round will neglect the impact of parameters with slightly lower change rates on personalization. 
%
%
%
The MaskUpgrade process in Algorithm \ref{alg2} ensures that parameters with significant changes are prioritized for personalization, while maintaining flexibility for other parameters with moderate changes.
In FedCPF, we comprehensively account for both the rate of parameter change and the number of training iterations.
Specifically, when determining the next set of personalized parameters $u_i^{r+1}$, we calculate the absolute value of the difference between each parameter before and after the update in the client model and average this difference over the number of updates to obtain the average parameter change rate.
%
We set a constant $\rm{acc}\%$, and each time the client model's accuracy reaches $k \cdot \rm{acc}\%$, where $k$ is a positive integer, the calculation range for the average change rate of the parameters $\Delta_{v_i^r}^{\rm{avg}}$ is adjusted.
%
%
For example, when $r = r_{\rm{th}}$, if the client model's accuracy first reaches $k \cdot \rm{acc}\%$, the calculation range for the average change rate of parameters in the current communication round $r_{\rm{cu}}$ will be $r \in [r_{\rm{th}}, r_{\rm{cu}}]$ until the accuracy reaches $(k+1) \cdot \rm{acc}\%$.
We define $\Delta_{v_i^r}^{\rm{avg}}$ as follows:
\begin{equation}
    \Delta_{v_i^r}^{\rm{avg}} = \frac {\sum_{r = r_{\rm{th}}}^{r_{\rm{cu}}} \Delta v_i^r} {r_{\rm{cu}} - r_{\rm{th}}},
\end{equation}
where $\Delta v_i^r = \vert v_i^{r+} - v_i^r \vert$ represents the absolute value of the difference before and after the parameter update.
Following the FedCPF training process outlined above, the parameters with the highest $p\%$ average change rate $\Delta_{v_i^r}^{\rm{avg}}$ are frozen as personalized parameters in the client model by setting the corresponding position in $\eta_i^r$ to 1.
%
%
By averaging the parameter change rate over multiple updates, we account for cumulative changes, ensuring that the most impactful parameters are selected for freezing.
This approach enables us to accurately select and freeze the personalized parameters within the client model, thereby enhancing the performance of the model within the personalized federated learning framework.

\begin{figure*}[t]
    \centering
    \includegraphics[clip,scale=0.23]{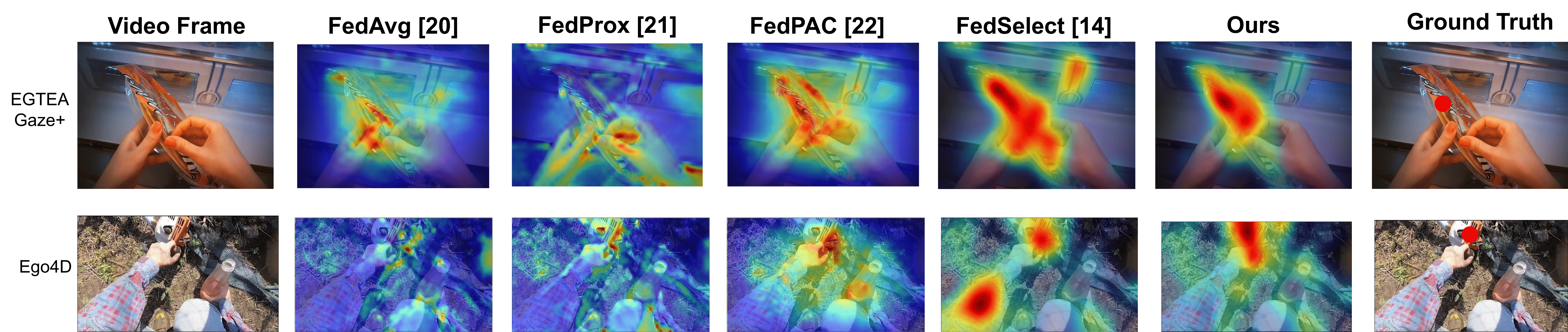}
    \vspace{-3mm}
    \caption{Qualitative evaluation result for gaze estimation. The predicted gaze is represented as a heatmap overlaid on input frames. In the Ground Truth image, the red dots refer to the actual gaze fixation records in the dataset.}
    \vspace{-5mm}
    \label{Qualitative Evaluation}
\end{figure*}

\vspace{-3mm}
\section{Experiments}
\vspace{-2mm}
\label{sec:experiments}
\subsection{Experimental Setting}
\vspace{-2mm}
\noindent \textbf{Dataset and Baseline.} We utilized the EGTEA Gaze+ \cite{li2018eye} and Ego4D \cite{grauman2022ego4d} datasets to verify the effectiveness of our proposed method. 
The EGTEA Gaze+ dataset includes 28 hours of egocentric video footage with gaze tracking annotations, while the Ego4D dataset consists of 27 videos, totaling 31 hours of gaze tracking data captured in diverse social environments.
The videos were segmented into 5-second clips to facilitate training and testing processes.
During training, the GLC module in each client was trained solely on its local data while continuously communicating with other clients to update the global model on the server. 
%
%
A total of 14,596 clips, comprising 6,456 from the EGTEA Gaze+ dataset and 8,140 from the Ego4D dataset, were used for training, while 3,439 clips, including 1,404 from EGTEA Gaze+ and 2,035 from Ego4D, were allocated for testing. These clips were extracted from 32 egocentric videos representing diverse sessions across the datasets.
Additionally, we employed a Vision Transformer (ViT \cite{dosovitskiy2020image}) with 8 blocks as the backbone architecture for FedCPF.

\noindent \textbf{Comparative Methods.} To assess the effectiveness of our approach, we compared it with several baseline methods. These include: 1) Local-only: Training the backbone model exclusively on each client dataset without any collaboration. 2) General Federated Learning Methods: Approaches like FedAvg \cite{konevcny2016federated} and FedProx \cite{li2020federated}, which aggregate client updates to train a shared global model. 3) Personalized Federated Learning (PFL) Methods: Techniques such as FedPAC \cite{xu2023personalized} and FedSelect \cite{tamirisa2024fedselect}, which focus on adapting models to individual clients.
%
In addition, we conducted an ablation study where the proposed framework was modified to randomly freeze parameters during the personalization process, rather than selecting them based on the average rate of change.
%
It is worth noting that the local-only results refer to the performance obtained by training a baseline model using only the local dataset of each client, without any collaboration or data sharing.


\noindent \textbf{Training Settings.} In our PFL framework, the number of participating clients was set to $N = 5$. 
The suppression matrix $\mathbf{S}$ was configured with $\lambda = 10^8$, ensuring that values in the first column and diagonal were mapped to probability distributions, while all other entries were effectively `masked out' upon applying the softmax function.
The number of communication rounds was fixed at $R = 100$ across all experiments.
The personalization rate $\rho$ was varied within the range $\{0.1, 0.2, 0.3, \dots, 0.7\}$, based on extensive experimental evidence indicating that the proposed method performs comparably across this interval.
The threshold $acc$, which adjusts the calculation range of the phased average rate of change, was tested across multiple values: $acc \in \{0.01, 0.05, 0.10, 0.15, 0.20\}$.
%

\noindent \textbf{Evaluation.} 
Recall, precision, and F1-score were used as evaluation metrics for binary gaze estimation, following prior work \cite{li2018eye, li2021eye} on EGTEA Gaze+ and Ego4D.
Aligning with the previous study \cite{feng2024privacy}, a saliency map was generated for each frame, and if the ground truth point was covered by a region with saliency values over 0.7, the prediction was considered successful. 
Recall was the ratio of successfully predicted frames, and precision was the ratio of accurate predictions within those frames. 
The final results were averaged across all test clips. 

\vspace{-3mm}
\subsection{Quantitative Evaluation}
\vspace{-2mm}
As shown in Table \ref{quantitative evaluation}, compared to the local-only approach, FedCPF demonstrates improved performance by leveraging communication between client models and the global model.
Comparisons with general federated learning methods reveal that our approach significantly enhances the personalization capabilities of client models by selectively freezing key parameters.
Furthermore, comparisons with other PFL methods, especially FedSelect, highlight the effectiveness of our proposed parameter selection strategy, which comprehensively accounts for the rate of parameter change.
Notably, the proposed method achieves measurable improvements over FedSelect, a state-of-the-art approach, emphasizing its practical significance. 
Additionally, a t-test on the EGTEA Gaze+ dataset indicated a statistically significant improvement in Recall for FedCPF (M = 55.85, SD = 0.74) compared to FedSelect (M = 54.69, SD = 1.16), with $\text{p-value} = 0.023$. 
A similar t-test for Precision on the same dataset showed a significant enhancement for FedCPF (M = 31.74, SD = 0.82) over FedSelect (M = 30.65, SD = 1.15), with $\text{p-value} = 0.025$.
On the Ego4D dataset, a t-test for Precision also demonstrated a statistically significant difference, where FedCPF (M = 25.2, SD = 0.92) outperformed FedSelect (M = 24.63, SD = 1.47), with $\text{p-value} = 0.045$. 
Similarly, a t-test for Recall showed a significant improvement for FedCPF (M = 37.10, SD = 0.28) outperformed FedSelect (M = 36.82, SD = 0.31), with $\text{p-value} = 0.05$. 
%
%
%
Therefore, although the observed improvements are modest, they highlight the robustness and practical applicability of the proposed framework.
%
%
%
Moreover, we conducted ablation experiments to validate the effectiveness of the parameter selection and freezing strategy in our proposed method.
FedCPF was trained under identical conditions ($\rho = 0.5, acc = 0.05$); however, during the personalization process, parameters were randomly selected and frozen instead of being chosen based on their average rate of change.
The selection process stopped once the personalization rate $\rho$ was reached, following the same procedure as the normal training workflow. 
The experimental results demonstrate that FedCPF (random freezing) underperforms the proposed method, confirming the efficacy of freezing parameters based on their average rate of change.
As shown in Table \ref{different personalization rate}, FedCPF achieves optimal performance when the personalization rate lies between 0.4 and 0.5.
The results in Table \ref{different accuracy threshold} indicate that the model attains its highest precision, recall, and F1-score when $acc$ is set to 0.05.
This threshold effectively balances the focus on historical parameter changes and current updates during the training process.
When $acc$ is smaller, the starting range for calculating the parameter change rate aligns more closely with the current round; conversely, a larger $acc$ shifts the starting range toward earlier rounds.
Experimental results suggest that the optimal balance is achieved when $acc$ equals 0.05.
%
%


\vspace{-3mm}
\subsection{Qualitative Evaluation}
\vspace{-2mm}
Figure \ref{Qualitative Evaluation} illustrates the qualitative comparison between the proposed method and other methods. 
It is evident from the figure that our approach predicts the user's gaze more accurately, achieving closer alignment with the Ground Truth. 
Notably, the proposed method demonstrates improved precision by predicting the correct region with a smaller, more accurate coverage area. 
These findings qualitatively validate the effectiveness of our PFL framework.

\vspace{-3mm}
\section{Conclusion}
\vspace{-3mm}
In this paper, we have introduced FedCPF, a novel framework for personalized federated learning in egocentric video gaze estimation with comprehensive parameter freezing. 
Rather than selecting personalized parameters based solely on the rate of change in a single round, we proposed a strategy that calculates the average rate of parameter change across multiple updates.
This comprehensive strategy enables more precise parameter selection, ensuring that the selected parameters are effectively frozen in client models for improved personalization. 
Furthermore, FedCPF's ability to maintain performance comparable to centralized models highlights its potential for real-world applications where data privacy is a critical concern.
The experimental results demonstrate the robustness and effectiveness of the proposed method, particularly its adaptability to diverse user data and its ability to preserve privacy in federated learning settings. These findings underscore its potential for real-world applications, such as AR/VR systems and assistive technologies.
%
%
%

\bibliographystyle{IEEEbib}
\bibliography{icip}

\end{document}